\documentclass{article}
\usepackage{lmodern}
\usepackage{amssymb,amsmath}
\usepackage{multicol}
\usepackage{ifxetex,ifluatex}
\ifnum 0\ifxetex 1\fi\ifluatex 1\fi=0 % if pdftex
  \usepackage[T1]{fontenc}
  \usepackage[utf8]{inputenc}
\else % if luatex or xelatex
  \ifxetex
    \usepackage{mathspec}
  \else
    \usepackage{fontspec}
  \fi
  \defaultfontfeatures{Ligatures=TeX,Scale=MatchLowercase}
\fi

\usepackage{lscape}
\usepackage{pgfplotstable} % Generates table from .csv

\usepackage[format=plain,
            labelfont=it,
            textfont=it]{caption}
\usepackage{url}
\usepackage{biblatex}
\IfFileExists{upquote.sty}{\usepackage{upquote}}{}
\IfFileExists{microtype.sty}{%
\usepackage{microtype}

\UseMicrotypeSet[protrusion]{basicmath} % disable protrusion for tt fonts
}{}

\usepackage[letterpaper, portrait, margin=1in]{geometry}

\usepackage{longtable,booktabs}
\usepackage{graphicx,grffile}

\makeatletter
\def\maxwidth{\ifdim\Gin@nat@width>\linewidth\linewidth\else\Gin@nat@width\fi}
\def\maxheight{\ifdim\Gin@nat@height>\textheight\textheight\else\Gin@nat@height\fi}
\makeatother
\setkeys{Gin}{width=\maxwidth,height=\maxheight,keepaspectratio}
\IfFileExists{parskip.sty}{%
\usepackage{parskip}
}{% else
\setlength{\parindent}{0pt}
\setlength{\parskip}{6pt plus 2pt minus 1pt}
}
\ifx\paragraph\undefined\else
\let\oldparagraph\paragraph
\renewcommand{\paragraph}[1]{\oldparagraph{#1}\mbox{}}
\fi
\ifx\subparagraph\undefined\else
\let\oldsubparagraph\subparagraph
\renewcommand{\subparagraph}[1]{\oldsubparagraph{#1}\mbox{}}
\fi
\usepackage[ruled]{algorithm2e}

\makeatletter
\@ifpackageloaded{subfig}{}{\usepackage{subfig}}
\@ifpackageloaded{caption}{}{\usepackage{caption}}
\AtBeginDocument{%

}

\@ifpackageloaded{float}{}{\usepackage{float}}
\floatstyle{ruled}
\@ifundefined{c@chapter}{\newfloat{codelisting}{h}{lop}}{\newfloat{codelisting}{h}{lop}[chapter]}
\floatname{codelisting}{Listing}

\makeatother

\title{A Massively-Parallel 3D Simulator for Soft and Hybrid Robots}
\usepackage{authblk}
\author[%
  1%
  ]{%
  Joel Clay%
}
\author[%
  1%
  ]{%
  Sofia Wyetzner%
}
\author[%
  1%
  ]{%
  Alex Gaudio%
}
\author[1]{Boxi Xia}
\author[1]{Andrew Moshova}
\author[1]{Jacob Austin}
\author[1]{Max Segan}
\author[%
  1%
  ]{%
  Hod Lipson%
}
\affil[1]{Columbia University, New York, NY, United States}
\date{}

\makeatletter
\def\@maketitle{%
  \newpage \null \vskip 2em
  \begin {center}%
    \let \footnote \thanks
         {\LARGE \@title \par}%
         \vskip 1.5em%
                {\large \lineskip .5em%
                  \begin {tabular}[t]{c}%
                    \@author
                  \end {tabular}\par}%
                                                \vskip 1em{\large \@date}%
  \end {center}%
  \par
  \vskip 1.5em}
\makeatother
\pgfplotsset{compat=1.15}
\begin{document}

\maketitle

\begin{figure}
    \centering
    \includegraphics{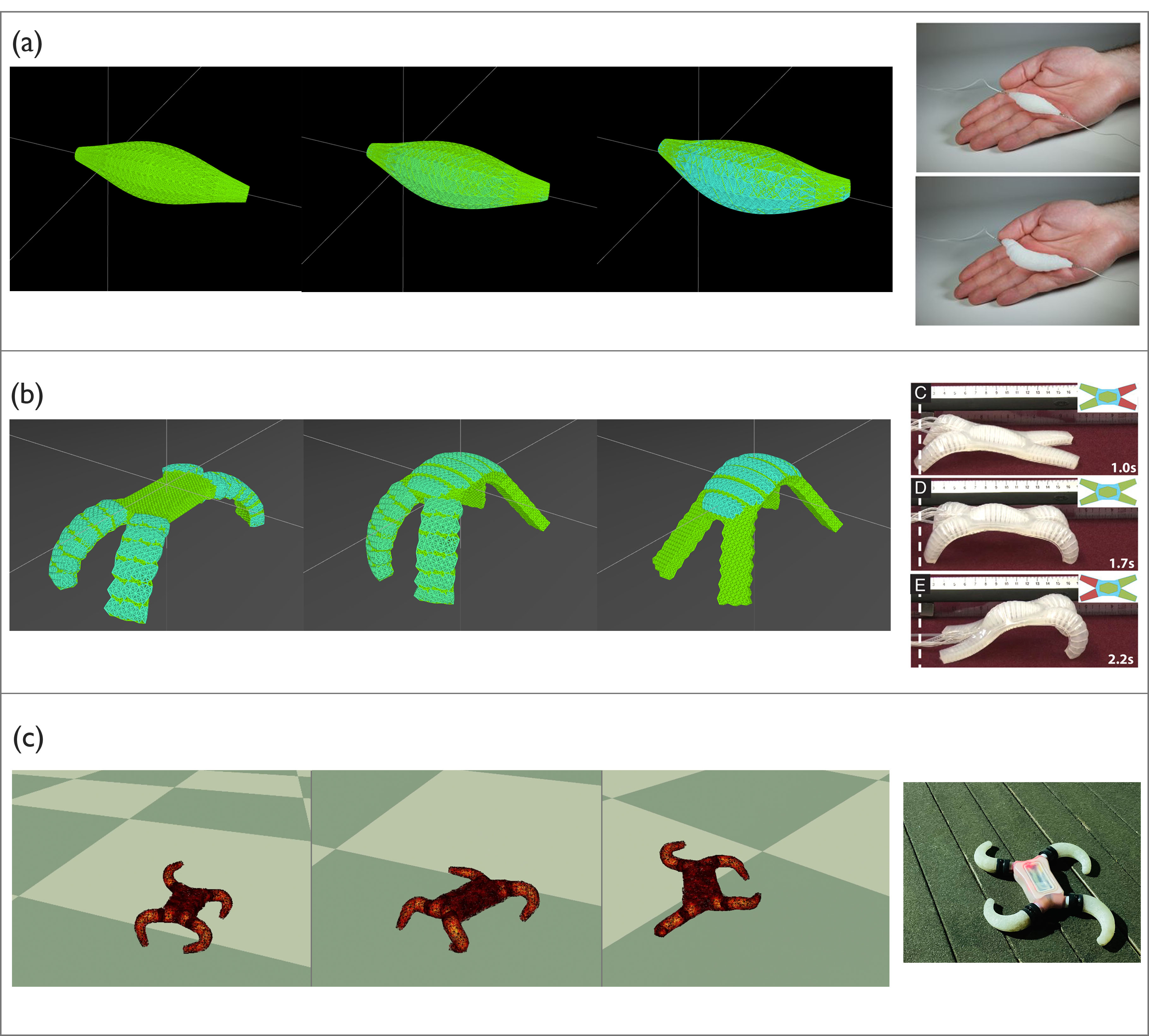}
    \caption{Images on the left contain frames from our method simulating actuation scenarios similar to the real examples on the right. (a) A simulation of the soft thermal actuator from \cite{miriyev_soft_2017} expanding. The actuated portion is slightly offset from the object's volume to demonstrate a similar pattern to the reference. The reference image on the right is modified from \cite{miriyev_soft_2017}. (b) A simulation of a walking soft robot based off of the seminal pneumatically-actuated soft robot in Shepherd et al.\cite{shepherd_multigait_2011}. The simulation is multi-material where actuated material creates periodic joints for limb deformation. The image on the right is modified from \cite{shepherd_multigait_2011} (c) A soft robot created by our group simulated by our method. The robot has four soft legs that are actuated by motors attached to their respective rotational joints creating fast locomotion. The simulation captures the resulting deformation and mobility on the left. The image on the right shows the physical robot.}\label{fig:walking_robot}
\end{figure}

  \subsection{Abstract}\label{abstract}
Simulation is an important step in robotics for creating control policies and testing various physical parameters. Soft robotics is a field that presents unique physical challenges for simulating its subjects due to the nonlinearity of deformable material components along with other innovative, and often complex, physical properties. Because of the computational cost of simulating soft and heterogeneous objects with traditional techniques, rigid robotics simulators are not well suited to simulating soft robots. Thus, many engineers must build their own one-off simulators tailored to their system, or use existing simulators with reduced performance. In order to facilitate the development of this exciting technology, this work presents an interactive-speed, accurate, and versatile simulator for a variety of types of soft robots. Cronos, our open-source 3D simulation engine, parallelizes a mass-spring  model for ultra-fast performance on both deformable and rigid objects. Our approach is applicable to a wide array of nonlinear material configurations, including high deformability, volumetric actuation, or heterogenous stiffness. This versatility provides the ability to mix materials and geometric components freely within a single robot simulation. By exploiting the flexibility and scalability of nonlinear Hookean mass-spring systems, this framework simulates soft and rigid objects via a highly parallel model for near real-time speed. We describe an efficient GPU/CUDA implementation, which we demonstrate to achieve computation of over 1 billion elements per second on consumer-grade GPU cards. Dynamic physical accuracy of the system is validated by comparing results to Euler--Bernoulli beam theory, natural frequency predictions, and empirical data of a soft structure under large deformation.

\subsection{Introduction}\label{introduction}

As soft robotics technologies improve and the range of soft materials expands, the ability to interactively model and simulate such systems must keep up \cite{lipson_challenges_2014}. Components of soft robots may be deformable, heterogeneous, self-actuated, or some combination thereof \cite{kim_soft_2013}. Traditional finite element physical simulators, while extremely successful at modeling linear physical systems, are challenged by simulating soft non-linear materials efficiently \cite{sifakis_fem_2012} especially when large deformations are involved. The need for adequate simulators for soft robotics is apparent for a variety of reasons, including for design validation, for topology optimization and generative design, and for training controllers in physically-realistic simulation. 
We propose that mass-spring systems provide a conceptually simple, inexpensive, and accurate solution that can elegantly handle the nonlinearity of highly deformable solids in a manner that is performant for the needs described above. We present the first open-source implementation of a parallelized mass-spring system that both can dynamically operate on upwards of 1 million springs in real time and can enable the simulation of objects with the complex physical properties of smart materials (Fig. \ref{fig:walking_robot}). Our simulator achieves massive parallelization via a relaxation approach based on techniques developed in the pre-computer era \cite{southwell_relaxation_1940}.

One facet of soft robotic fabrication that is both challenging and increasingly relevant is the use of multiple interspersed materials within a single component \cite{lipson_challenges_2014}. In order to model a heterogeneous material, the simulator must handle both soft and stiff material within one object. This ability to simulate rigid materials with the same method as deformable materials has generally been bounded by performance. Rigid bodies alone can be simulated with traditional Finite Element Methods efficiently and with high accuracy, but simulations require custom methods for nonlinear elastic materials and actuatable materials. In particular, finite element approaches are difficult to parallelize across multiple processors, and therefore do not scale well to massively parallel multicore systems, where most of the price performance gains have been achieved in the past decade \cite{keckler_gpus_2011}.

Mass-spring systems are traditionally a simple and fast method for simulating deformable bodies, but become unstable for materials with high Young's moduli using higher time steps. However, we propose that with a sufficiently small time step, mass-spring systems can handle rigid and deformable components. Optimizing for performance allows us to accommodate materials with very low elasticity in addition to soft materials by using a small time step while also achieving interactive real-time simulation.

Our simulation technique successfully models both rigid and deformable high-resolution objects in interactive real-time. With functional accuracy and subject flexibility, our engine additionally supports high-resolution topologies and interspersed material properties. This opens up possibilities for simulating and fabricating complex soft components quickly, which were previously limited to longer processes. By comparing our simulator to existing systems, we show our simulator’s potential for interactive design an design automation \cite{hiller_automatic_2011} as the first open-source verified-accuracy mass-spring simulation engine.

% Graphics
\subsubsection{Deformable/Rigid Body Modeling Methods}
There is a large body work addressing simulation of deformable materials. The computer graphics community has created methods over the past several decades for physics-based visual simulations of soft objects \cite{nealen_physically_2006}. The ability to use large time steps is valued in computer graphics, as scenes can be complex and need to be rendered in real-time, often without any failure, such as in video game engines. While these motivations are slightly different from the aims of the engineering community, to whom this paper is tailored, the goal of visual accuracy has produced many robust physics-based techniques, which we will discuss here.

There are several high-performant techniques created for graphic simulation of deformable bodies. Liu et al., \cite{liu_fast_2013}, presents an implicit solver for mass-spring systems that is faster than Newton's method but not inherently parallelizable. Constraint-based techniques offer a numerically robust alternative to mass-spring models for simulating soft objects. Position Based Dynamics, a popular technique for visually simulating deformations, uses constraint projection on positions rather than force accumulation and integration \cite{muller_position_2007}. Particles update their positions unconstrained, which are then corrected in order to fit elasticity constraints. The method uses a Gauss-Seidel-type solver, and therefore is not easily parallelizable. This method was extended by Bouaziz et al. to incorporate aspects of the Finite Element Method, which improves accuracy and robustness. It uses a Jacobi-type solver for increased (but not full) parallelism \cite{bouaziz_projective_2014}. However it has been shown that these methods are not suited to stiff objects or objects with non-constant material properties \cite{tournier_stable_2015}, as they were intended specifically for deformable isotropic solids.

\subsubsection{Multi-Materials and Self-Actuation}
Few of these methods were created with the intent of simulating multi-material or self-actuated objects. Traditional Finite Element Methods have been shown to become inaccurate for heterogeneous materials, requiring computationally-expensive custom methods to be developed \cite{ten_thije_large_2007}. There has been work on simulating animated characters made of a stiff "skeleton" and a soft surrounding "flesh" using Projective Dynamics \cite{li_fast_2019}. However this technique does not seem to accommodate a spectrum of material stiffness, instead focusing on a rigid-soft binary. Huang et al. presents a numerical simulation method for specifically for actuated soft robotics using second-order integration on a discrete elastic rod model \cite{huang_dynamic_2020}. Their method accurately captures the kinematics of several types of common soft robotics structures, however the method appears to be limited in the realm of larger or multi-material systems. On the other hand, mass-spring models can be implemented with the flexibility to handle a variety of stiffnesses within a single system.

Traditional mass-spring systems have been notably used in graphics to simulate 1D hair \cite{selle_mass_2008}, 2D cloth \cite{provot_deformation_1995}, and 3D volumetric objects \cite{liu_fast_2013}. In non-visual application, mass-spring methods are often used to simulate soft deformable bodies for robotics applications in surgery simulation, due to their suitability for modeling deformations in complex organs composed of different tissue types \cite{meier_real-time_2005, patete_multi-tissue_2013, wang_vascular_2015, etheredge_parallel_2011, marchal_towards_2008}. Mass-spring systems are capable of handling multiple materials via their inherent discretization of springs. In addition we show that self-actuation can be enabled by dynamically modifying volumetric properties.

\subsubsection{Simulation Software for Soft Robots}
Software packages such as Nvidia PhysX and Nvidia FLeX offer pre-packaged GPU-accelerated implementations of physics-derived graphics algorithms to model physcial phenomena \cite{noauthor_physx_2018, noauthor_nvidia_2015}. Specifically, both pieces of software rely on a Position Based dynamics approach for deformable bodies \cite{rieffel_evolving_2009, mrowca_flexible_2018}. These implementations are either confined to rigid structures or derived from constraint-based physics, therefore limiting their accurate simulation to a smaller range of soft mechanisms.

In soft robotics, there are existing simulation engines intended for simulating physically-accurate components. Our work philosophically builds off of the Voxelyze library \cite{hiller_dynamic_2014}, which decomposes 3D objects into voxels with discrete material properties, using a mass-spring-like model for simulation. For contrast, our work here does support a voxel-based lattice, but we demonstrate support for more complex geometries as well. Coevoet et al. presents soft robotic simulation software that uses FEM techniques to achieve high physical accuracy for a variety of fabricated soft robots \cite{coevoet_software_2017}. However these techniques were not specifically designed to achieve real-time performance, especially for more complex discretizations. Hu et al. presents the ChainQueen simulator, a soft-robotics simulation engine that supports many features of novel soft robots, including self-actuation, and heterogenous materials using the Material-Point Method \cite{hu_chainqueen:_2019}. The primary focus of ChainQueen is differentiability, which is enabled through our system but not the focal point. Thus there is a space for a fast and usable implementation that can handle a wide array of structures and material properties, for which we provide our solution.

\subsection{Method}\label{method}

In this section, we review the mass-spring model our implementation utilizes, along with the three integration methods we use for performance and accuracy comparison.

\begin{figure}
    \centering
    \includegraphics[height=150 pt]{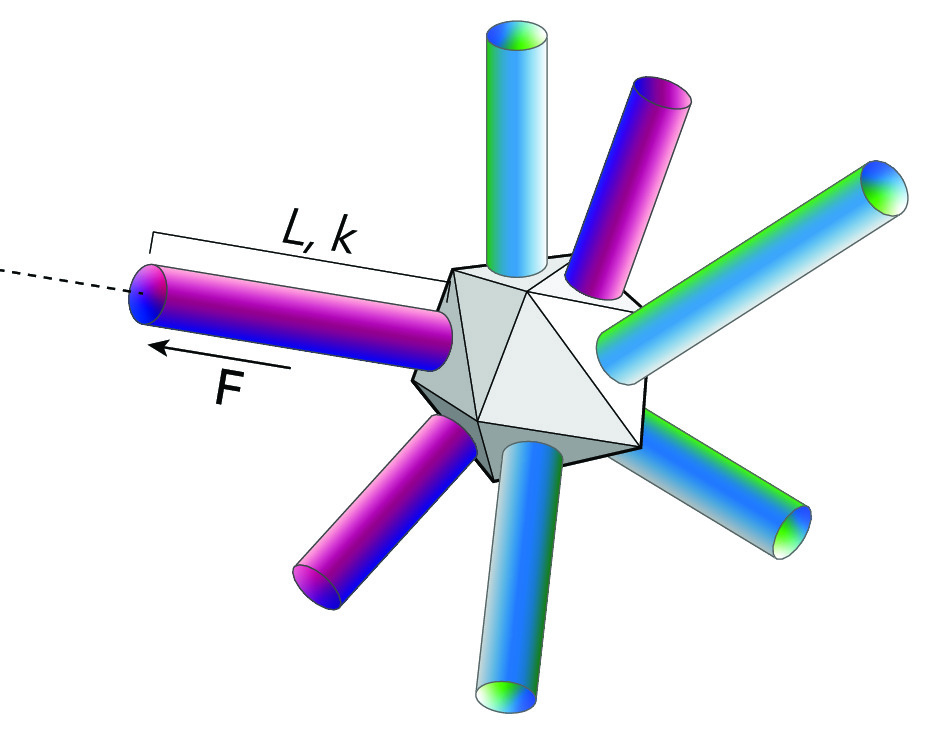}
    \caption{A diagram of a mass and attached springs with irregular geometry. Each spring has a defined rest length, $L$, and a Hookean spring constant, $k$, which apply a force, $F$, to connected masses. Our system allows for minute control over the structural and material properties of the simulated lattice, allowing springs to be individually rigid, actuated, or functionally parameterized.}\label{fig:random_zoom}
\end{figure}

 \subsubsection{Mass-Spring Force Distribution}\label{mass-spring-systems}

Mass-spring systems using Hookean springs distribute forces at each simulation iteration according to Newton's second law (Fig. \ref{fig:random_zoom}). These updates can be defined as follows. 

The state of each of the
\(N\) masses in the system is defined by their position
\(\vec{x}_i(t)\), mass \(m_i\) and the total force
\(\vec{f}_i(t, \vec{x}_i(t))\), \(0 \leq i \leq N-1\). Each mass \(m_i\)
may be connected to another mass \(m_j\) via a spring \(S_{ij}\). (In
practice, note that \(m_i\) typically has at most \(C_i\) spring
connections with \(C_i \leq 26\).) The acceleration
\(\vec{a}_i(t) = \frac{d^2 \vec{x}_i(t)}{dt^2}\) of mass \(m_i\) is
related to the total force \(\vec{f}_i\) via Newton's second law,
\begin{equation}
  m_i \vec{a}_i(t) = \vec{f}_i(t, \vec{x}_i(t)),
  \quad 0 \leq i \leq N-1.
\end{equation}

The mass-spring model decomposes \(\vec{f}_i\) as the sum of three
forces, \begin{equation}
  \vec{f}_i(t, \vec{x}_i(t)) = m_i \vec{g} + \vec{f}^{ext}_i(t)
  + \sum_{j=1}^{C_i} k_{ij}(|\vec{l}_{ij}(t)| - l_{ij}^0) \frac{\vec{l}_{ij}(t)}{|\vec{l}_{ij}(t)|},
\end{equation} where \(m_i \vec{g}\) is the gravitational force,
\(\vec{f}^{ext}_i(t)\) the external force and the third term is Hooke's
Law with \(\vec{l}_{ij}(t) = \vec{x}_j(t) - \vec{x}_i(t)\), rest length
\(l_{ij}^0\) and spring constant \(k_{ij}\).

The final system to integrate is \begin{equation}
  \frac{d}{dt}
  \begin{bmatrix}
    \vec{x}_i(t) \\
    \vec{v}_i(t)
  \end{bmatrix}
  =
  \begin{bmatrix}
    \vec{v}_i(t) \\
    \frac{1}{m_i}\vec{f}_i(t, \vec{x}_i(t))
  \end{bmatrix},
  \quad 0 \leq i \leq N-1,
\end{equation} where \(\vec{v}_i\) denotes the velocity of \(m_i\), and
for given initial position and velocity \(\vec{x}_i(0)\) and
\(\vec{v}_i(0)\).

At this point, a time integration method is used to update positions to the next time iteration.

  \subsubsection{Time Integration Schemes}\label{time-integration}

Our method uses several explicit time integration schemes in conjunction with the previously-described Hookean force distributions. Explicit time integration methods were chosen over implicit methods for their computational cheapness. The following offers a review of these methods.

We discretize time with a uniform time-step, \(\Delta t\), and write
\(t_n = n \Delta t\), \(n \geq 0\).

\textit{Forward Euler} 

The formula for this first-order method is for
\(0 \leq i \leq N-1\), \begin{equation}
  \begin{bmatrix}
    \vec{x}_i(t_{n+1}) \\
    \vec{v}_i(t_{n+1})
  \end{bmatrix}
  =
  \begin{bmatrix}
    \vec{x}_i(t_{n}) \\
    \vec{v}_i(t_{n})
  \end{bmatrix}
  +
  \Delta t
  \begin{bmatrix}
    \vec{v}_i(t_n) \\
    \frac{1}{m_i}\vec{f}_i(t_n, \vec{x}_i(t_n))
  \end{bmatrix},
  \quad n \geq 0.
\end{equation}

\textit{Verlet} 

Verlet integration is a second-order method that has
improved stability and accuracy over forward Euler at no extra
performance cost. The formula is given for \(0 \leq i \leq N-1\) by
\begin{align}
  \vec{x}_i(t_{1})
   & =
  \vec{x}_i(0)
  +
  \Delta t \vec{v}_i(0)
  +
  \frac{\Delta t^2}{2 m_i} \vec{f}_i(0, \vec{x}_i(0)),
  \\
  \vec{x}_i(t_{n+1})
   & =
  2\vec{x}_i(t_{n})
  -
  \vec{x}_i(t_{n-1})
  +
  \frac{\Delta t^2}{m_i} \vec{f}_i(t_{n}, \vec{x}_i(t_n)),
  \quad n \geq 1.
\end{align}

\textit{RK4} 

We have also provided an implementation using the fourth
order Runge--Kutta method \cite{press_numerical_2007}. In practice, we find the algorithm to be
stable, but performance suffers drastically since it requires the
evaluation at four different stages. Those stages are defined by
\begin{align}
  \vec{a}^n_i     & = \vec{v}_i(t_n),
  \quad
                  & \vec{A}^n_i                         & = \frac{1}{m_i}\vec{f}_i(t_n, \vec{x}_i(t_n)),                              \\
  \vec{b}^n_i     & = \vec{v}_i(t_n) + \vec{a}^n_i/2,
  \quad
                  & \vec{B}^n_i                         & = \frac{1}{m_i}\vec{f}_i(t_n + \Delta t/2, \vec{x}_i(t_n) + \vec{A}^n_i/2), \\
  \vec{c}^{\,n}_i & = \vec{v}_i(t_n) + \vec{b}^n_i/2,
  \quad
                  & \vec{C}^n_i                         & = \frac{1}{m_i}\vec{f}_i(t_n + \Delta t/2, \vec{x}_i(t_n) + \vec{B}^n_i/2), \\
  \vec{d}^{\,n}_i & = \vec{v}_i(t_n) + \vec{c}^{\,n}_i,
  \quad
                  & \vec{D}^n_i                         & = \frac{1}{m_i}\vec{f}_i(t_n + \Delta t, \vec{x}_i(t_n) + \vec{C}^n_i),
\end{align} and the formula for the update is

\begin{equation}
  \begin{bmatrix}
    \vec{x}_i(t_{n+1}) \\
    \vec{v}_i(t_{n+1})
  \end{bmatrix}
  =
  \begin{bmatrix}
    \vec{x}_i(t_{n}) \\
    \vec{v}_i(t_{n})
  \end{bmatrix}
  +
  \frac{\Delta t}{6}
  \begin{bmatrix}
    \vec{a}^n_i + 2\vec{b}^n_i + 2\vec{c}^{\,n}_i +\vec{d}^n_i    \\
    \vec{A}^n_i + 2{\vec{B}}^n_i + 2{\vec{C}}^n_i + {\vec{D}}^n_i \\
  \end{bmatrix}.
\end{equation}

  \subsubsection{Generating a Lattice}\label{generating-a-lattice}

The use of a mass-spring model relies upon discretizing a 3D space into a lattice mesh of nodes and edges (Fig. \ref{fig:femur_lattice}). There are several ways to achieve this, and here we present the two that we used in our testing: (1) Generating a cubic lattice by breaking up the space with a voxel grid; (2) Generating a quasi-uniform lattice via targeted random point selection.

\begin{figure}
    \centering
    \includegraphics{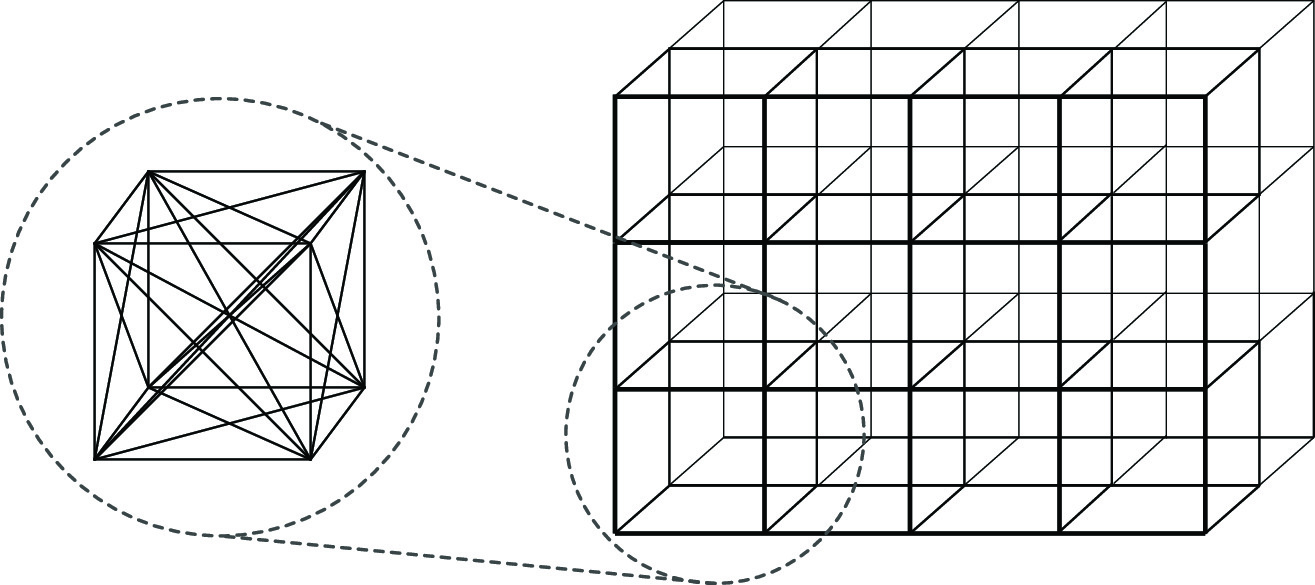}
    \caption{For a voxelized lattice, we build a grid of spring masses by first setting the mass
      positions, then linking each with neighboring masses to form a cube. We
      also link the face diagonals (duplicate faces are merged), and long
      diagonals.}\label{fig:figure1}
  
\end{figure}

For the former, a voxel grid of desired resolution is overlaid on the object space (most often defined by a 3D mesh file), and nodes which are not inside of the triangle-grid of the object boundaries are culled. Edges are generated between neighboring nodes (Fig. \ref{fig:figure1}). The resolution of this lattice is confined to arrangements describable in voxel form, but the resulting form is straightforward, providing clarity for testing and for multi-material simulations.

For the latter, we aim to sample the 3D space within a bound while maintaining a minimum distance between points. In order to achieve this Poisson disk spacing, we apply a method based on Mitchell's best candidate algorithm \cite{mitchell_spectrally_1991}. We begin by placing a single randomly chosen
vertex. The following vertex is chosen by generating 100 random
candidate vertices, and selecting the vertex which maximizes the
distances from its \(k\)-nearest neighbors. This is repeated until the
desired density of the lattice is achieved. This algorithm yields a
quasi-uniform lattice and ensures the vertices are appropriately spaced. We have demonstrated a quasi-uniform mass-spring lattice generated from a 3D mesh of a human femur (Fig. \ref{fig:femur_lattice}).

Once the lattice as been defined, masses are generated from nodes and springs from edges. Physical material properties derive spring constants and mass values. Mass values may be calculated by material density over the tetrahedral volume containing the point mass, or by using a uniform mass distribution based on the total mass of the structure. Spring constants can be derived through material elasticity moduli. Note that a full derivation of spring constants from physical properties is outside of the scope of this paper, but the spring constant often depends the resolution of the lattice in addition to the material elasticity. We have found that a starting point of 0.1 kg per mass with 10,000 as the default spring constant (scaling inversely to the length of the spring) creates a deformable material that can be simulated in real time with a time step of $0.0001$. Masses may have external force applied on them, and constraints such as planes might be configured. Finally, the resulting mass-spring lattice is copied to GPU memory to initialize simulation.

% Discuss material property derivation (density, mass, k)

% Discuss damping, time step, etc.

\begin{figure}
    \centering
    \includegraphics{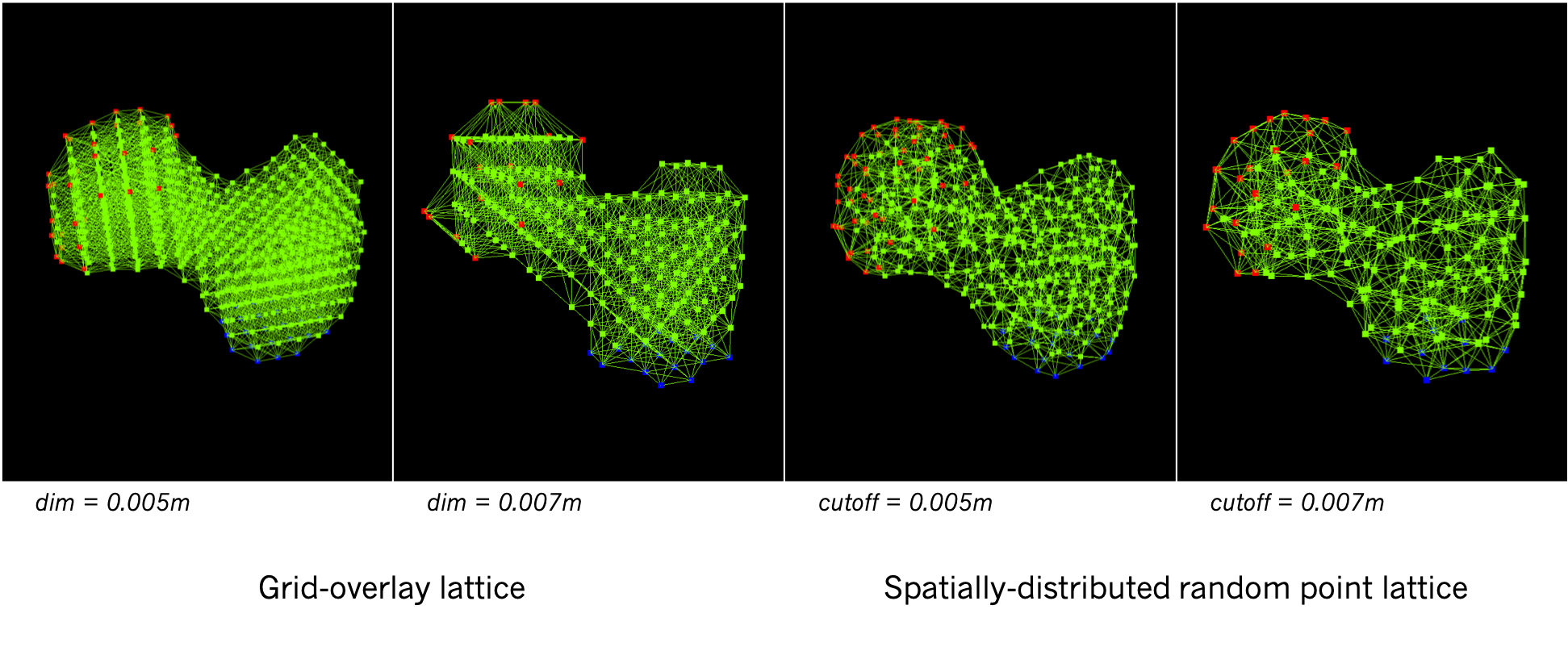}
    \caption{Mass-spring lattices generated from the 3D geometry of a human femur. On the left are two examples of a voxel-based lattice with different resolutions (voxel edge dimension is indicated by "dim"). On the right are two examples of random quasi-uniform lattices generated by a best candidate algorithm with different resolutions (minimum distance is indicated by "cutoff"). }\label{fig:femur_lattice}
 
\end{figure}

  \paragraph{GPU Implementation}\label{data-structures}

The springs and masses are represented internally by classes with member
variables to hold necessary state. We store the spring and masses
consecutively in memory so that access amongst threads remains fast, and
it also allows for faster operations when syncing from CPU to GPU or
vice versa.

We duplicate the storage of spring-to-mass connection data since there are
cases where using either the GPU or CPU is more suitable for performance. As we store the springs sequentially in memory, the masses can be accessed by iterating
through the springs and accessing their endpoints. Our parallel
implementation requires that we have the connected springs
stored on the mass, so that when visiting each mass, we can iterate
through the connected springs and perform the updates accordingly.

  \subsubsection{Parallel Algorithm}\label{parallel-implementation}

In a naive parallel-update version of the mass-spring system where masses are accessed freely from their connected springs, race conditions arise if the state of a mass is read or written to by multiple threads at once. One way to solve this is
by using atomic operations when updating masses. In the typical case of
each mass having more than 20 spring connections, these atomic updates
can often lead to a situation where threads applying the spring forces
will be competing for read/write access on the connected masses.

Depending on the structure of the masses, we have found this to lead to
a substantial reduction in throughput. This can be especially
problematic with CUDA since the order in which threads are dispatched is undefined, eliminating the strategy of ordering the thread dispatches to
minimize duplicate mass updates \cite{noauthor_programming_nodate}.

We have used a modified algorithm from [19]. The original algorithm solves the race condition by moving atomic operations and adding well-timed thread synchronizations. For double-precision vector operations, we have found that atomic operations lead to
reduced throughput due to the need for competing threads to wait on the
double-precision operations to complete. In testing, we have found double-precision
operations to be up to 10x slower on CUDA than their floating point
alternatives.

In order to avoid atomic operations on doubles, the springs are processed first to
append their forces to the connected masses. The array of forces on each mass are summed and applied in a later update step. This results in only needing an atomic operation
on the integer representing the index to insert into the array. We found
this to yield a substantial increase in throughput, allowing our engine to
exceed the 1 billion springs / second barrier.

\begin{algorithm}[H]
  \DontPrintSemicolon
  \SetKwData{F}{$\vec{f}$}
  \SetKwData{Fext}{$\vec{f}_{ext}$}
  \ForAll{springs $S_{ij} $}{
  calculate spring length $l_{ij} \leftarrow \sqrt{(\vec{x}_j(t) - \vec{x}_i(t))^2}$\;
  calculate spring force $\F_{spring} \leftarrow k_{ij} (l_{ij} - l_{ij}^0) \frac{\vec{x}_j(t) - \vec{x}_i(t)}{l_{ij}}$\;
  $m_i \leftarrow masses[i]$ \\
  $m_j \leftarrow masses[j]$ \\
  $fidx_1 \leftarrow$ atomicAdd($m_i$.force\_idx, 1) \\
  $fidx_2 \leftarrow$ atomicAdd($m_j$.force\_idx, 1)\;
  $m_i$.forces[$fidx_1$] $\leftarrow \F{spring}$ \\
  $m_j$.forces[$fidx_2$] $\leftarrow -\F{spring}$ \\
  }
  synchronize threads\;
  \ForAll{masses $M_i$}{
  \For{$i = 0, i < M_i$.force\_idx, $i++$} {
  $M_i$.force $\leftarrow M_i$.forces[$i$]\;
  }
  \tcp{Loop over force\_indices and sum into force on mass.}
  \tcp{Perform position/velocity integration}
  }
  \caption{Parallel Simulation Process}
\end{algorithm}

  \subsection{Results}\label{validation}

In order to validate that the mass-spring system approximates physical systems accurately, we have performed four tests: (1) a Euler--Bernoulli
cantilever beam test, (2) a physical accuracy test based on real material behavior, (3) a total energy test to ensure that energy in the system remains constant, and (4) an analytic natural frequency comparison.

  \subsubsection{Euler--Bernoulli Beam
    Tests}\label{eulerbernoulli-beam-tests}

In order to validate the accuracy of the simulation, standard Euler--Bernoulli beam theory was used to measure the first
fundamental frequency of a beam with length \(L\), height \(H\) and
width \(W\) to test against the frequency calculated from the output simulation positions of the same geometry.

The equation for the first fundamental frequency of a horizontal
cantilever beam subject to free vibration and a uniformly distributed
load (gravity) is, \[
  f = \frac{K}{2\pi} \sqrt{\frac{EIg}{\rho A L^4}},
\] where \(K\) is a constant corresponding to the first fundamental
frequency, \(E\) is the modulus of elasticity, \(I = H^3W/12\) is the
second moment of inertia along the \(y\)-axis, \(g\) is the
gravitational constant, \(\rho\) is the mass density, and \(A = WH\) is
the area of a \(yz\) cross-section. Simplifying the formula and treating
\(K\), \(E\), \(\rho\), and \(g\) as constants yields
\[f \propto \sqrt{\frac{H^2}{L^4}}.\] Therefore, for varying width, we
expect \(f\) to be a constant, for varying height, we expect \(f\) to be
affine, and for varying length, we expect \(f\) to decay quadratically.

A small load is applied to the end of the beam to create a deflection. This is on the order of a fraction of a millimeter, because the Euler-Bernoulli theory most accurately predicts small deflections.

The approach is as follows:

\begin{enumerate}
  \def\labelenumi{\arabic{enumi}.}
  \item
        Apply a small load to the tip of the beam.
  \item
        Begin the simulation with a small amount of simple velocity damping
        (0.01\%).
  \item
        Run the simulation for a set interval (500 ms), until the beam has relaxed.
  \item
        Release the damping and trace the position of the tip over time as
        the beam vibrates.
  \item
        Determine frequency with a zero-cross count method.
\end{enumerate}
\vspace{20 pt}

A tip mass of the cantilever beam was chosen in simulation to measure y-position for a period of 1000 simulation milliseconds. The starting y-position of the traced mass is recorded, and the number of times that the the mass passes through this point is counted. Once the number of times that the mass crosses through its starting point is determined within a time interval, the frequency of the beam is calculated.

\begin{figure}
    \centering
    \includegraphics{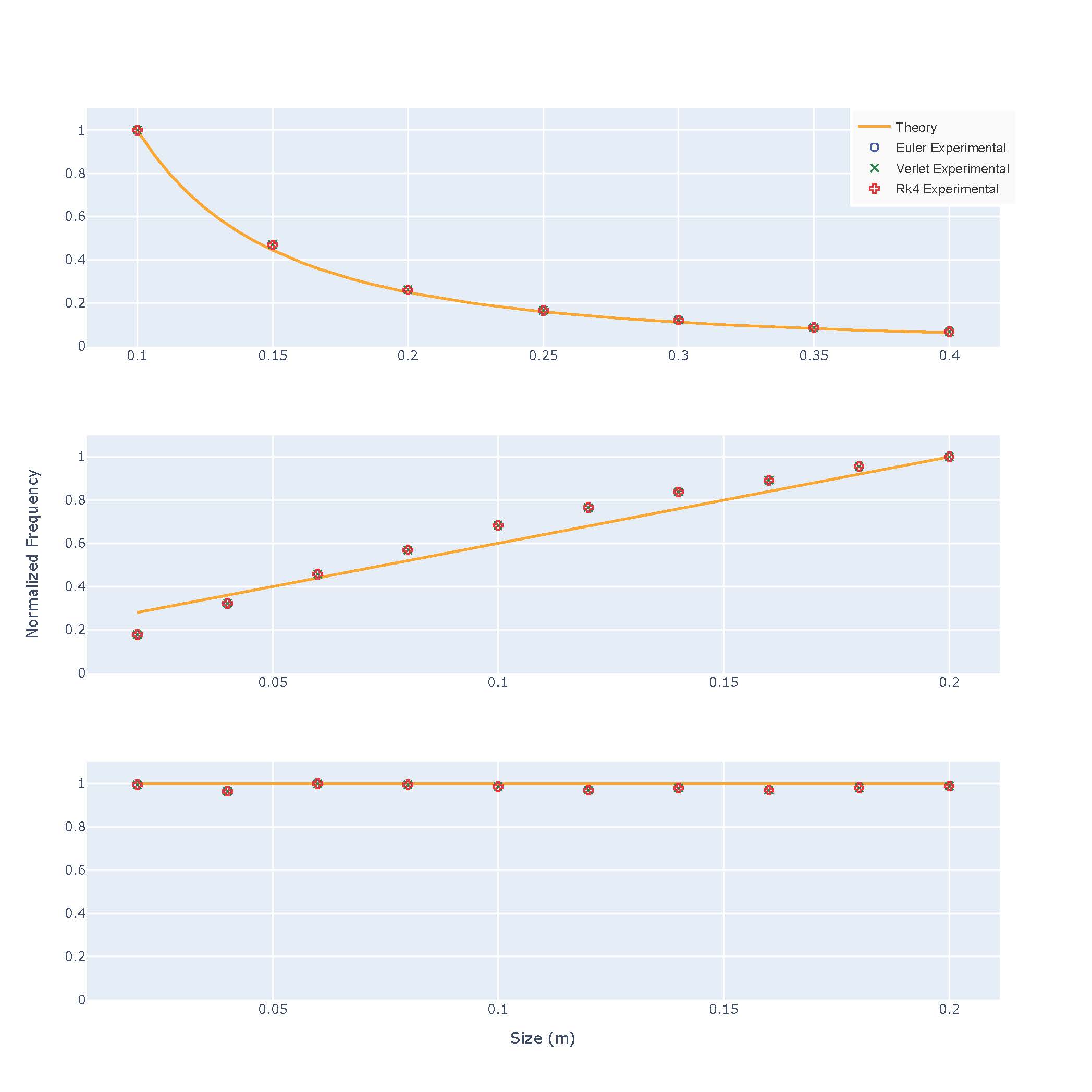}
    \caption{We show normalized frequencies as a function of change in
      length (top), width (middle) and height (bottom) for the three different
      timestepping schemes. The output of the normalized frequencies
      demonstrate that our mass-spring method is a close approximation to
      expected theoretical values.}\label{fig:rk4_analysis}
  
\end{figure}

Below, (Fig. \ref{fig:rk4_analysis}), we have shown the results of these experiments being
run on a variety of beam sizes by independently varying height, width, and
length, and calculating the resulting frequency. We compared these with analytical results to determine how closely our mass-spring system
follows the Euler--Bernoulli beam theory.

Based on the results, we find that the mass-spring cantilever beam is
often a close approximation of what we find in Euler--Bernoulli beam
theory. We find that for varying widths, the experimental results are largely greater than the beam theory predicted frequencies. We are not able to determine whether the theory or the experimental results is a better predictor of the physical phenomenon, but the uniqueness of this discrepancy in the Y direction is notable.

\pagebreak

\subsubsection{Accuracy for Large Deformations}
\begin{figure}
    \centering
    \includegraphics[width=0.75\textwidth]{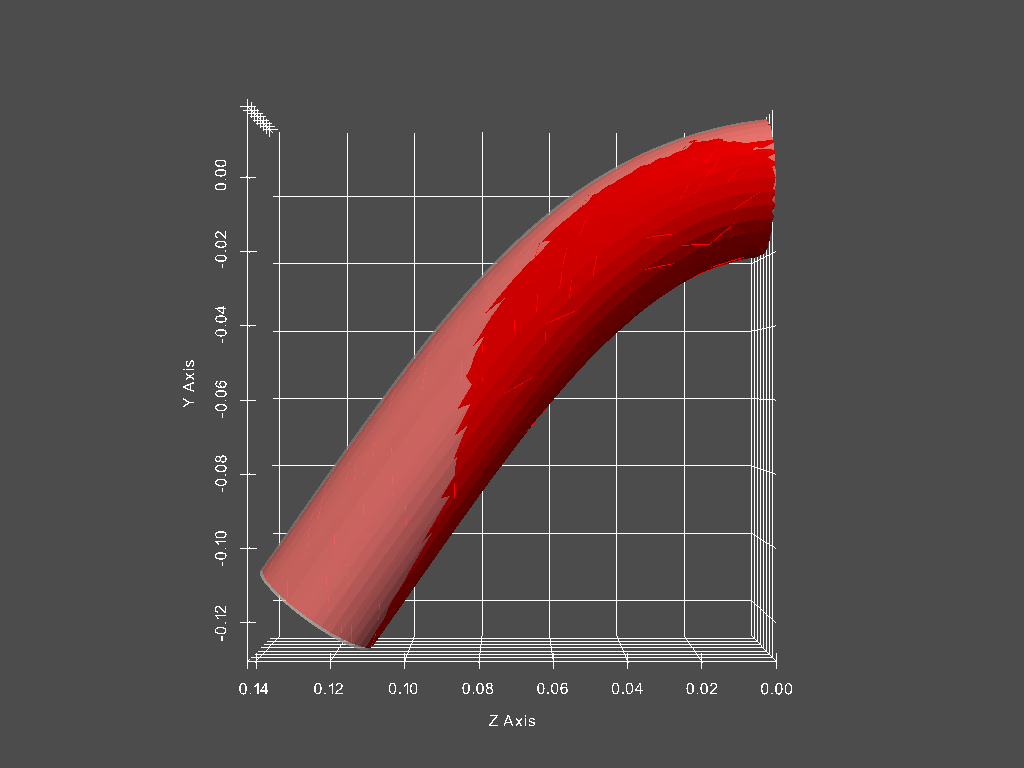}
    \caption{We show accuracy validation of our method for a soft cylinder that has a large deflection applied. The empirical reference is overlaid in white and is from \cite{marchal_towards_2008}. The result of the deflected cylinder simulated by our method is displayed in red.}\label{fig:largeDeflection}
  
\end{figure}

Marchal et al. presents us with an excellent data set for validating large deformations centered around a scan of a physical deformable cylinder along with several traditional simulation methods including three FEM techniques \cite{marchal_towards_2008}. We have constructed the beam with the parameters described in their paper using our simulation method. The two beams are shown overlapping in Fig. \ref{fig:largeDeflection}. We then perform a surface analysis on the beam relaxed by our method against the empirical beam, whose 3D models can be found as included in the SOFA simulator library \cite{noauthor_sofa_nodate}.

\begin{longtable}[]{@{}llllll@{}}
  \caption{\label{tbl:large_def_table} We compare our method's accuracy on large deformations against finite element techniques.}\tabularnewline
\toprule
  \begin{minipage}[b]{0.49\columnwidth}\raggedright
    Method\strut
  \end{minipage} & \begin{minipage}[b]{0.29\columnwidth}\raggedright
    Relative Surface Error (mm)\strut
  \end{minipage}\tabularnewline
  \midrule
  \endhead
  \begin{minipage}[t]{0.49\columnwidth}\raggedright
    Mass-Spring [ours]\strut
  \end{minipage} & \begin{minipage}[t]{0.29\columnwidth}\raggedright
    1.68\strut
  \end{minipage}\tabularnewline
  \begin{minipage}[t]{0.49\columnwidth}\raggedright
    Mass-Spring [Marchal et al.]\strut
  \end{minipage} & \begin{minipage}[t]{0.29\columnwidth}\raggedright
    0.75\strut
  \end{minipage}\tabularnewline
  \begin{minipage}[t]{0.49\columnwidth}\raggedright
    Linear FEM Tetrahedral [Marchal et al.]\strut
  \end{minipage} & \begin{minipage}[t]{0.29\columnwidth}\raggedright
    18.60\strut
  \end{minipage}\tabularnewline
  \begin{minipage}[t]{0.49\columnwidth}\raggedright
    Co-rotational FEM Tetrahedral [Marchal et al.]\strut
  \end{minipage} & \begin{minipage}[t]{0.29\columnwidth}\raggedright
    0.63\strut
  \end{minipage}\tabularnewline
  \begin{minipage}[t]{0.49\columnwidth}\raggedright
    Co-rotational FEM Hexahedral [Marchal et al.]\strut
  \end{minipage} & \begin{minipage}[t]{0.29\columnwidth}\raggedright
    2.87\strut
  \end{minipage}\tabularnewline
  \bottomrule
\end{longtable}

Table \ref{tbl:large_def_table} presents a direct comparison between our method and several simulated methods. The empirical beam was used for surface comparison. The co-rotational tetrahedral FEM and listed mass-spring method have less surface error than our method, but the values are comparable. Our method was tuned according to visual accuracy, as in the mass-spring model in Marchal et al. Thus the surface error our method presents could be potentially lower for slightly better parameter values. Our method shows a high level of accuracy, which beats the linear tetrahedral and co-rotational hexahedral FEM surface errors, as tested by Marchal et al. According to the authors' conclusions, these methods overpredict the deformation significantly.

  \subsubsection{Energy Conservation Tests}\label{total-energy-tests}

We have performed energy tests that consist of calculating the elastic
potential energy, gravitational potential energy, and
kinetic energy of each spring. With this simple validation, we have shown that the mass-spring system obeys the law of conservation of energy. The energy tests were performed in tandem with the beam tests above via the Euler integration scheme.

\begin{figure}
    \centering
    \includegraphics{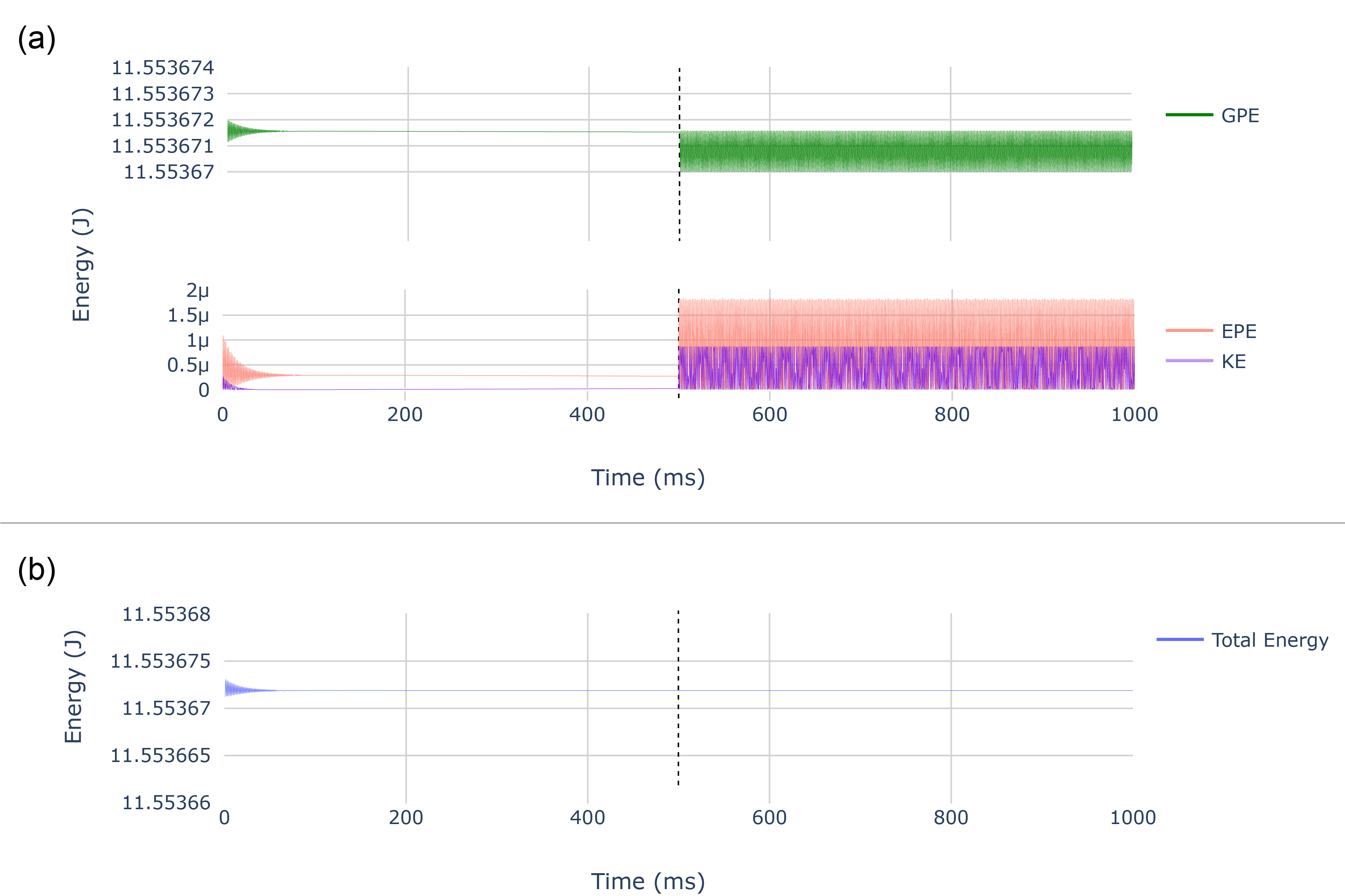}
    \caption{Total energy of the system shows energy conservation after damping is released. (a) demonstrates the components of total energy: EPE = elastic potential energy, GPE = gravitational potential energy, KE = kinetic energy. (b) is a close up view of the total energy sum of (a). The vertical gray lines at 500 ms indicate when the load is released from the beam and damping is set to 0.}\label{fig:energy_tests}
  
\end{figure}

The results of this test can be seen in (Fig. \ref{fig:energy_tests}). As expected the system briefly oscillates around an energy value during relaxation while damping is applied, as shown in the beginning of the graphs. In addition, we can see the predicted oscillation of gravitational potential energy, kinetic energy, elastic potential energy during the second half of the test. This indicates the movement of the tip of the beam and resolves to a constant total energy.

\subsubsection{Natural Frequency Analysis}

\begin{figure}
    \centering
    \includegraphics[width=4in]{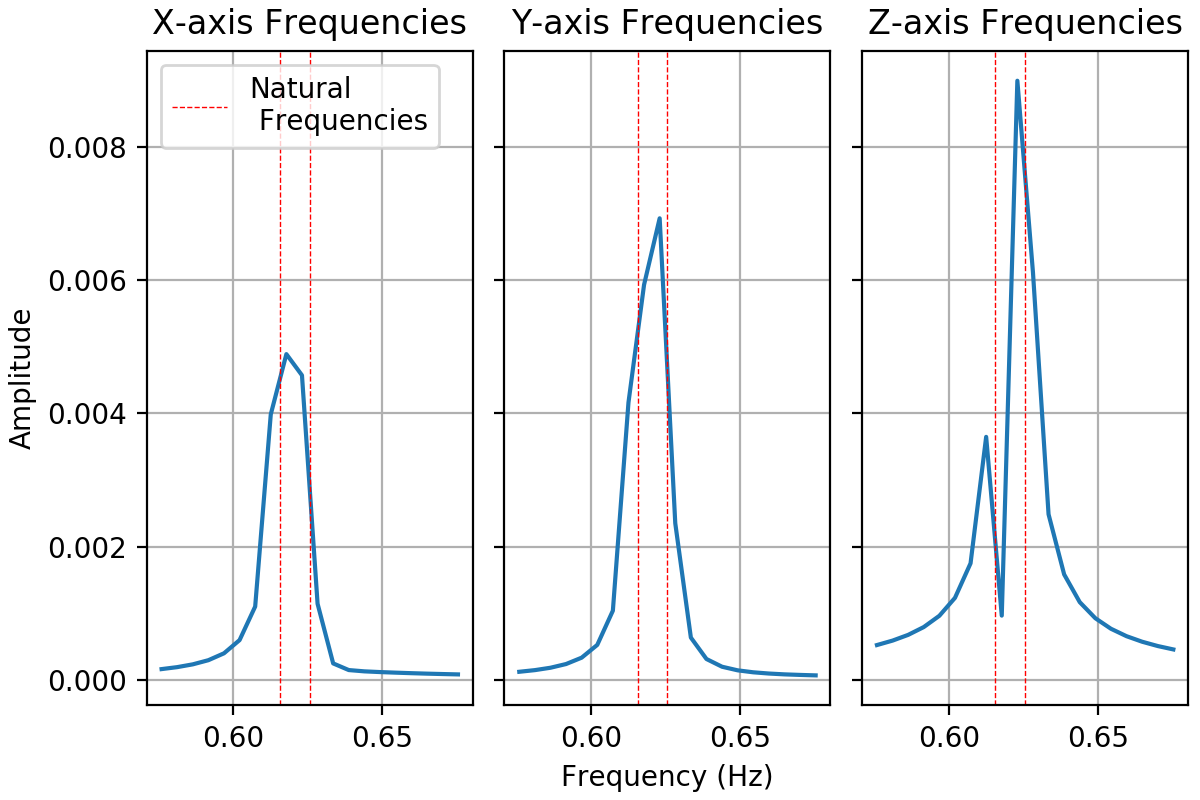}
    \includegraphics{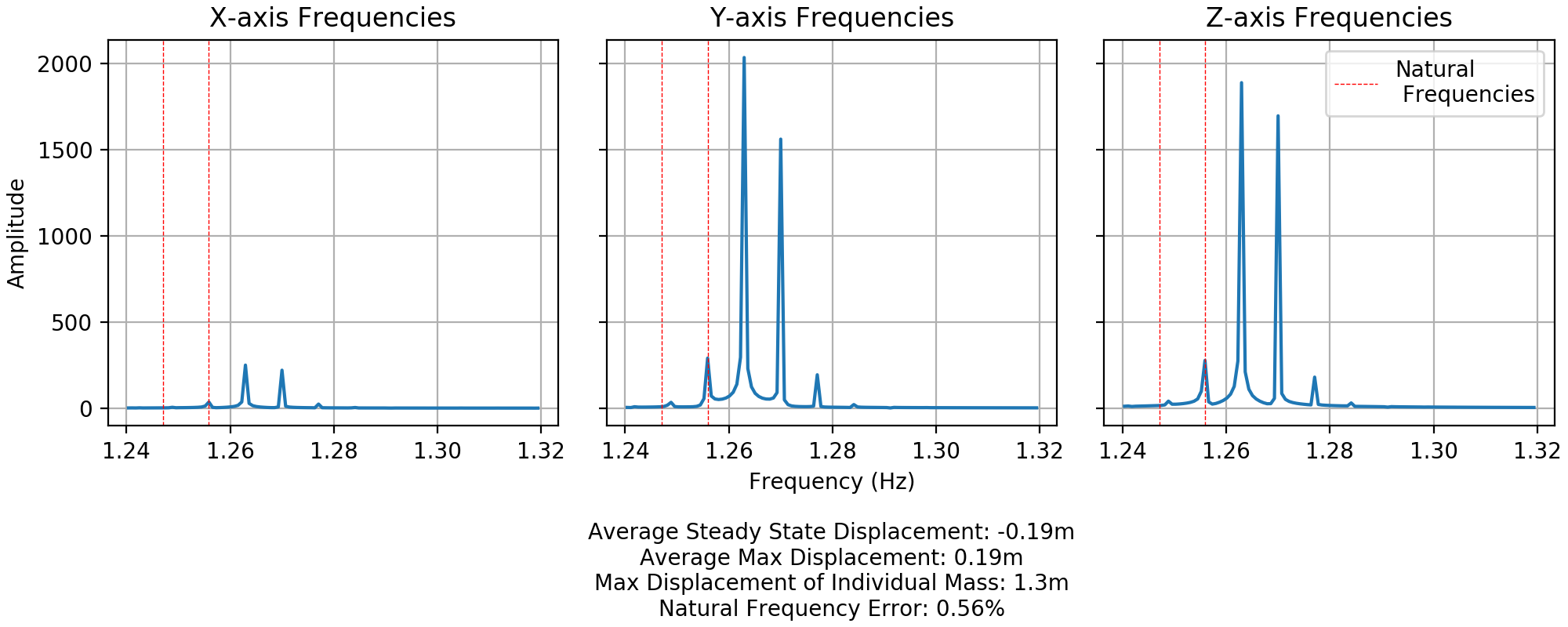}

    \caption{Graphs showing natural frequency prediction on a 10m cantilever beam. The red dotted lines indicate the analytical predictions, with the blue indicating the result of the simulated model under FFT. (Top) Frequency prediction on a cantilever beam under small deformation. As demonstrated, the frequencies are predicted with a high level of accuracy. (Bottom) Frequency prediction on a cantilever beam under large deformation. }
    \label{fig:natfreq}
\end{figure}

To further our validation, we can analytically predict natural frequencies of simulated objects during model runtime. The natural frequency of a system is determined by solving the generalized eigenvalue problem. We construct a mass matrix and a stiffness matrix and then use a sparse Cholesky decomposition solver to calculate the two smallest frequencies. Fig. \ref{fig:natfreq} shows these predicted frequencies in comparison to the actual behavior of the model. Modeled frequency is measured by tracking mass displacement, with the final values calculated via a Fast Fourier Transform. As demonstrated, the behavior of the system closely mirrors the analytical predictions for small deformations. We a discrepancy of 0.56\% error for a cantilever beam that has undergone a large deformation, and the resulting graphs have been magnified for clarity. We suspect that this discrepancy is caused by the analytical solution's limits in approximating large deformations, and that the model achieves a behavior closer to reality.

  \subsubsection{Performance}\label{performance}

Below are the results of a performance analysis on our implementation. The analysis was performed on several GPUs and one CPU to demonstrate the speed up due to parallelization. Here, we prove the high efficiency and capacity that this implementation achieves.

\begin{figure}
    \centering
    \includegraphics{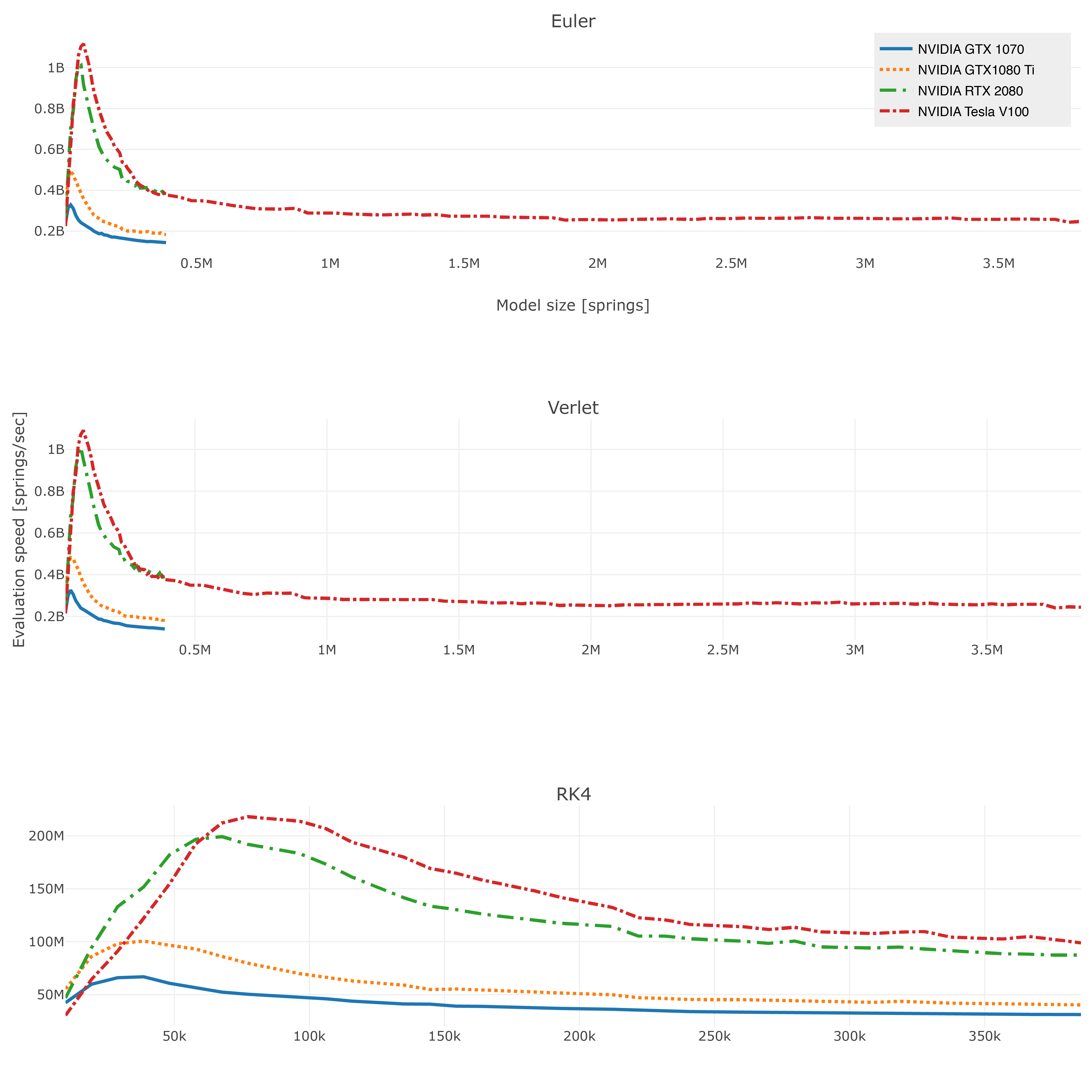}
    \caption{We compare the performance profiles across different
      integration methods and GPU cards. Euler and Verlet exhibit similar
      characteristics, while RK4 suffers from dramatically reduced
      throughput.}\label{fig:perf_analysis}
  
\end{figure}

\begin{longtable}[]{@{}llllll@{}}
  \caption{\label{tbl:perf_table}We compare the peak performance for an
    array of computing devices. We achieve over 1B springs per second on
    Nvidia's V100 and 2080 RTX, while performance drops significantly for
    older devices. Further optimizations are possible, as shown in the preliminary results on the RTX 3090.}\tabularnewline

  \toprule
  \begin{minipage}[b]{0.19\columnwidth}\raggedright
    Device\strut
  \end{minipage} & \begin{minipage}[b]{0.09\columnwidth}\raggedright
    Bars\strut
  \end{minipage} & \begin{minipage}[b]{0.11\columnwidth}\raggedright
    Bars/Sec\strut
  \end{minipage} & \begin{minipage}[b]{0.10\columnwidth}\raggedright
    Cores\strut
  \end{minipage} & \begin{minipage}[b]{0.17\columnwidth}\raggedright
    Max Clock (GHz)\strut
  \end{minipage}\tabularnewline
  \midrule
  \endhead
  \begin{minipage}[t]{0.19\columnwidth}\raggedright
    Nvidia Tesla V100\strut
  \end{minipage} & \begin{minipage}[t]{0.09\columnwidth}\raggedright
    77.19k\strut
  \end{minipage} & \begin{minipage}[t]{0.11\columnwidth}\raggedright
    1.12B\strut
  \end{minipage} & \begin{minipage}[t]{0.10\columnwidth}\raggedright
    5120\strut
  \end{minipage} & \begin{minipage}[t]{0.17\columnwidth}\raggedright
    1.530\strut
  \end{minipage}\tabularnewline
  \begin{minipage}[t]{0.19\columnwidth}\raggedright
    Nvidia RTX 2080\strut
  \end{minipage} & \begin{minipage}[t]{0.09\columnwidth}\raggedright
    67.55k\strut
  \end{minipage} & \begin{minipage}[t]{0.11\columnwidth}\raggedright
    1.01B\strut
  \end{minipage} & \begin{minipage}[t]{0.10\columnwidth}\raggedright
    2944\strut
  \end{minipage} & \begin{minipage}[t]{0.17\columnwidth}\raggedright
    1.710\strut
  \end{minipage}\tabularnewline
  \begin{minipage}[t]{0.19\columnwidth}\raggedright
    Nvidia GTX 1080 Ti\strut
  \end{minipage} & \begin{minipage}[t]{0.09\columnwidth}\raggedright
    28.99k\strut
  \end{minipage} & \begin{minipage}[t]{0.11\columnwidth}\raggedright
    492.34M\strut
  \end{minipage} & \begin{minipage}[t]{0.10\columnwidth}\raggedright
    3584\strut
  \end{minipage} & \begin{minipage}[t]{0.17\columnwidth}\raggedright
    1.582\strut
  \end{minipage}\tabularnewline
  \begin{minipage}[t]{0.19\columnwidth}\raggedright
    Nvidia GTX 1070\strut
  \end{minipage} & \begin{minipage}[t]{0.09\columnwidth}\raggedright
    28.99k\strut
  \end{minipage} & \begin{minipage}[t]{0.11\columnwidth}\raggedright
    328.11M\strut
  \end{minipage} & \begin{minipage}[t]{0.10\columnwidth}\raggedright
    1920\strut
  \end{minipage} & \begin{minipage}[t]{0.17\columnwidth}\raggedright
    1.683\strut
  \end{minipage}\tabularnewline
  \begin{minipage}[t]{0.19\columnwidth}\raggedright
    Intel i7-8700k\strut
  \end{minipage} & \begin{minipage}[t]{0.09\columnwidth}\raggedright
    9.72k\strut
  \end{minipage} & \begin{minipage}[t]{0.11\columnwidth}\raggedright
    23.38M\strut
  \end{minipage} & \begin{minipage}[t]{0.10\columnwidth}\raggedright
    6\strut
  \end{minipage} & \begin{minipage}[t]{0.17\columnwidth}\raggedright
    4.700\strut
  \end{minipage}\tabularnewline
    \begin{minipage}[t]{0.19\columnwidth}\raggedright
    Nvidia RTX 3090\footnotemark{}\strut
  \end{minipage} & \begin{minipage}[t]{0.09\columnwidth}\raggedright
    60.84k\strut
  \end{minipage} & \begin{minipage}[t]{0.11\columnwidth}\raggedright
    9.11B\strut
  \end{minipage} & \begin{minipage}[t]{0.10\columnwidth}\raggedright
    10496\strut
  \end{minipage} & \begin{minipage}[t]{0.17\columnwidth}\raggedright
    1.695\strut
  \end{minipage}\tabularnewline
  \bottomrule
\end{longtable}

\footnotetext{A further enhancement of the algorithm was used on this device demonstrating further performance improvements, primarily driven by the removal of explicit locking for synchronization.}

For larger structure sizes, we find that the CUDA implementation yields a nearly 10x improvement in throughput. We notice a performance decrease
as the number of masses exceeds the cores on the GPU. This is likely due to the overhead incurred by CUDA's dispatcher waiting for running threads to finish before dispatching additional threads. If the GPU core count continues growing at the same speed, we predict that the GPU performance will continue to significantly outpace that of the CPU.

In Fig. \ref{fig:perf_analysis} we demonstrate peak performance on various GPUs. Note that except for the Nvidia Tesla V100, the cards are challenged by loading large model sizes in RAM, and model sizes drop off at around 0.4M springs as a result. We naturally find that peak performance for these GPUs is achieved at far fewer springs than the this limit.

In order to benchmark our simulator against readily available alternatives, we compared performance results based on model iteration time. The data from alternative approaches was replicated from Hu et al \cite{hu_chainqueen:_2019}. We used a similar test case as the given example with 8,000 active masses and 25 anchored (inactive) masses on the Nvidia GTX 1080 Ti, the GPU used in the Hu et al. benchmarking. Our approach out-performs Nvidia Flex and ChainQueen’s forward propagation (the more similar operation to our step, and also faster than the respective backwards propagation operation in ChainQueen). 

\begin{table}
\caption{\label{tbl:comp_table}We compare our performance against other approaches to soft robot simulation software. Note that while the term "particles" accurately describes the discretization technique for both Nvidia Flex and ChainQueen, the best analog in our system are mass elements. "FP" stands for the Forward Propagation operation in ChainQueen as opposed to Backwards Propagation.}\tabularnewline
\begin{tabular}{@{}lllll@{}}
\toprule
Simulator    & Particles   & Model Iteration Time \\ \midrule
Nvidia Flex         & 8,024       & 3.5 ms \\
ChainQueen (FP)   & 8,000    & 0.392 ms \\
Ours    & 8,000 (active)    & 0.216 ms \\
\bottomrule
\end{tabular}%
\end{table}

  \subsubsection{Multi-GPU}\label{multi-gpu}

In these performance tests, we have not considered  multi-GPU
workstations. Keeping the same level of throughput with Multiple-GPU
CUDA support is made difficult by the fact that memory operations must
still pass through the PCI-E interface. Due to the way in which the
forces propagate throughout the entire structure, we require the changes
made on one GPU to be synced to the other GPUs before the next time
step. This memory copy operation between GPUs is costly to the
throughput as the current PCIe cards can only achieve throughput of up
to 32 GB/s. With the recent introduction of NVLink on data center cards,
speeds of up to 300 GB/s can be realized \cite{noauthor_programming_nodate}. The increase in
inter-GPU throughput represents an opportunity to leverage larger GPU
core counts. The 10x increase in throughput from NVLink could minimize
time waiting for GPU memory copies and enable our parallel
implementation to better scale across multiple-GPU configurations.

  \subsubsection{Applications of the Simulation Method}\label{applications}
  
 In Fig. \ref{fig:walking_robot} we present several examples of our simulator handling soft robotic components. Visualizations are done through an analogous implementation system of ours, which was built on the same implementation scheme but is slightly slower due to a focus on graphics support \cite{austin_titan:_2019}. The pneumatically actuated bending robot shown is based on the real-world robot presented in Shepherd et al. \cite{shepherd_multigait_2011}. There are five sets of striped actuators, four for each legs and one in the main body, that are activated independently. Notable similar motion is achieved as the physical analog by Sheperd et al.; the motion is shown in our attached supplemental video. In addition the soft-legged locomotive robot developed by our group contains a mix of two materials: a soft material for the legs and a rigid material for the body. We visually tuned material and actuation parameters to achieve likeness to the physical robots in the respective examples. These applications demonstrate our method handling multi-material objects, functional expressions of spring properties, and actuation.

\subsection{Conclusion}\label{conclusion}

We have introduced the first open-source, massively-parallel, mass-spring simulation engine with verified performance and accuracy. Our implementation of a GPU mass-spring system simulates soft objects with minutely customizable material properties at near real-time speed for interactive simulation and optimization.

Our engine was tested with three different integration schemes for accuracy and performance. On state-of-the-art GPUs our system demonstrates great speed, processing 1 billion spring updates per second. In addition our approach exploits the computational cheapness of mass-spring systems to out-perform alternative soft robotics platforms. 

A series of cantilever beam experiments were also run with the software to compare the results to the analytical solutions. In addition we validated our simulator against available physical data of soft materials. We have found the parallel mass-spring method to be a good approximation for simulating deformable and rigid structures. 

We have used an interactive GUI on top of this engine to create and modify objects containing over 1 million springs in real time. Figures \ref{fig:walking_robot}, \ref{fig:femur_lattice}, and supplemental videos were generated via this GUI, which contains the slightly slower version of this method mentioned above that was created for visualization purposes. Our method can also be used to successfully model objects containing multiple materials with varying stiffness and objects with actuated components. Our system is implemented in C++ with the CUDA runtime library and includes Python bindings for further usability. The code is fully open-source, and we hope it will be used to model a wide variety of complex components in soft robotics.

  \subsection{Acknowledgments}\label{acknowledgments}

This work has been funded in part by U.S. Defense Advanced Research Project Agency (DARPA) TRADES grant number HR0011-17-2-0014 and by Israel Ministry of Defense (IMOD) grant number 4440729085 for Soft Robotics.

An open-source library implementing this simulator is available online \cite{noauthor_titan_nodate, austin_titan:_2019}. Source code for the software used specifically to perform the experiments described in this paper is available is available at \cite{noauthor_creative_nodate}.

\subsection{Appendix}
\subsubsection{A. Open Source Library}
A fully open source library, Titan, that is actively developed and maintained is available for installation and download at \cite{noauthor_titan_nodate}. This library was used for the visualizations of this method, and it uses the OpenGL API to render the results of the engine in real time. Titan was written in C++, but Python bindings are also available for usability. Titan contains functionality for discretizing soft robots into mass-spring networks and then simulating the bodies with a user-defined time resolution. Constraints may be specified on portions of the robots, which may be directional or planar. Contact planes can be configured to apply frictional forces to robots for simulating locomotion.

\pagebreak

\printbibliography

\end{document}